\documentclass[letterpaper, 10 pt, conference]{ieeeconf}  

\IEEEoverridecommandlockouts                             

\usepackage{graphicx}
\usepackage{cite}
\usepackage{amsmath,amssymb,amsfonts}
\usepackage{algorithmic}
\usepackage{graphicx}
\usepackage{textcomp}
\usepackage{threeparttable}
\usepackage{pifont}
\usepackage{color}
\usepackage[table]{xcolor}

\usepackage{CJKutf8}

\usepackage[T1]{fontenc}
\usepackage{gensymb}
\usepackage{adjustbox}
\usepackage{etoolbox}
\usepackage{array}
\usepackage{booktabs}
\usepackage{multirow}
\usepackage{dsfont}
\usepackage{makecell}

\title{\LARGE \bf
A three-dimensional force estimation method for the cable-driven soft robot based on monocular images
}

\author{Xiaohan Zhu$^{\#}$, Ran Bu$^{\#}$, Zhen Li, Fan Xu$^{*}$ and Hesheng Wang$^{*}$ 
\thanks{This work was supported in part by the Natural Science Foundation of China under Grant 62203298. Corresponding Author: Fan Xu, Hesheng Wang.} 
\thanks{X. H. Zhu, Z. Li, F. Xu and H. S. Wang are with School of Electronic Information and Electrical Engineering, the Department of Automation, Shanghai Jiao Tong University (email: xufanlyra@sjtu.edu.cn and wanghesheng@sjtu.edu.cn). R. Bu is with Artificial Intelligence Research Institute and School of Information and Control Engineering, China University of Mining and Technology.} 
\thanks{\# These authors have contributed equally to this work.}
}

\begin{document}

\begin{CJK}{UTF8}{gbsn}

\maketitle
\thispagestyle{empty}
\pagestyle{empty}

\begin{abstract}

Soft manipulators are known for their superiority in coping with high-safety-demanding interaction tasks, e.g., robot-assisted surgeries, elderly caring, etc. Yet the challenges residing in real-time contact feedback have hindered further applications in precise manipulation. 
This paper proposes an end-to-end network to estimate the 3D contact force of the soft robot, with the aim of enhancing its capabilities in interactive tasks. The presented method features directly utilizing monocular images fused with multidimensional actuation information as the network inputs. This approach simplifies the preprocessing of raw data compared to related studies that utilize 3D shape information for network inputs, consequently reducing configuration reconstruction errors. 
The unified feature representation module is devised to elevate low-dimensional features from the system's actuation signals to the same level as image features, facilitating smoother integration of multimodal information. 
The proposed method has been experimentally validated in the soft robot testbed, achieving satisfying accuracy in 3D force estimation (with a mean relative error of 0.84\% compared to the best-reported result of 2.2\% in the related works).

\end{abstract}

\section{INTRODUCTION}



It is important to obtain an accurate knowledge of the interaction force between the soft robot end effector and the environment in tasks such as surgery, exploration of unstructured environments, and object manipulation. Existing studies leverage direct-measuring methods by embedding strain gauge, MEMS-based, piezoelectric, optical fiber Bragg sensors into soft manipulators\cite{uzun2020optical}, proving high accuracy in certain tasks. Though showing potential in addressing force feedback of soft mechanism, such sensors are most uncommercialized and may require high-precision or high-cost manufacturing process \cite{nazari2021image}, hindering their integration into soft robots, especially at small scale. 
Indirect force-sensing technology has also gained wide attention recently\cite{diezinger20223d}. Such methods mainly lay the basic principle of the intrinsic relationship between the deflection of soft robots and external forces, making it possible to estimate forces with known shapes.




According to\cite{black2017parallel}, classical approaches to indirect force sensing of soft robots can be divided into two categories: actuation-based\cite{yasin2021joint} and deflection-based\cite{aloi2019estimating} methods. 
Actuation-based methods use joint-level information such as cable tension\cite{hao2023two}, actuation position\cite{black2017parallel} or actuation forces\cite{xu2010intrinsic} to calculate the external forces, the performance of which is sensitive to joint-level friction, hysteresis and modeling uncertainties. 
Deflection-based methods emphasize accurate configuration measurements using, for instance, fiber Bragg gratings(FBGs)\cite{khan2017force} and vision system\cite{diezinger20223d}. These two types of methods both rely on the known system model based on the Cosserat rod theory \cite{back2015catheter} or piece-wise constant curvature assumption \cite{hasanzadeh2015model} to compute external forces. 
Earlier study\cite{diezinger20223d} utilized a stereo-vision system to reconstruct the 3D shape of a flexible rod. They estimated the tip force by solving force and moment equilibrium equations, achieving a satisfying accuracy of 2.2\%.
Compared to stereo camera systems, monocular camera systems enjoy higher economy efficiency, lower requirements for room size, and easier deployment thus more preferable in applications. A recent study \cite{brumfiel2024image} using a monocular camera to obtain shape measurements has also been investigated. To the best of the author's knowledge, this paper represents the initial attempt to reconstruct the configuration of a soft robot using a monocular camera and introduces a deflection-based force estimation algorithm.

\begin{figure}[!t]
\centerline{\includegraphics[width=\columnwidth]{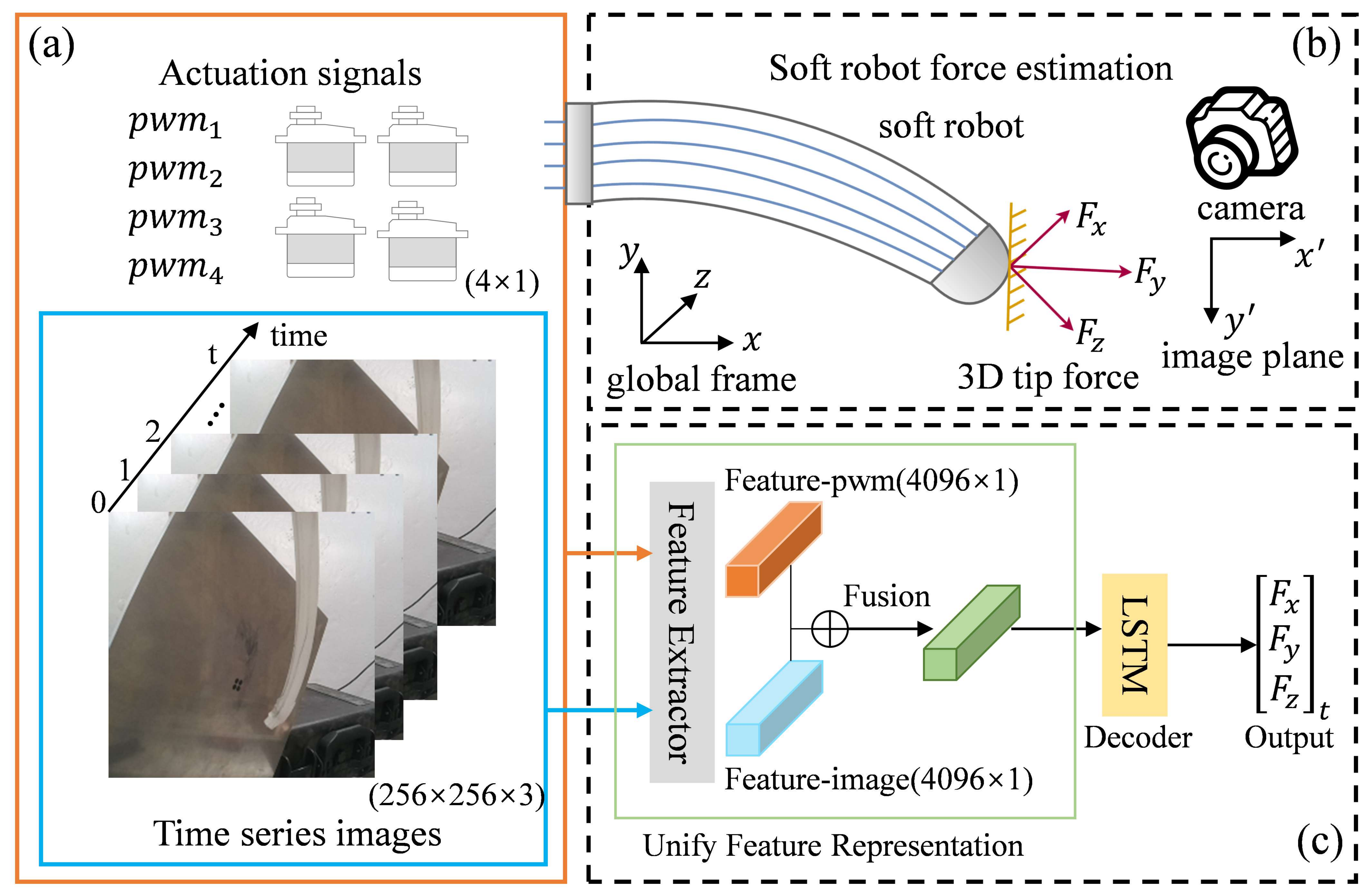}}
\vspace{-3mm}
\caption{ (a) Inputs of the proposed end-to-end network are monocular image sequences and actuation signals. (b) Schematic diagram of soft robot 3D tip force estimation. (c) The main process of the proposed method. }
\label{introduction_fig}
\vspace{-7mm}
\label{fig0}
\end{figure}


Deep learning methods provide an alternative besides the model-based methods, showing promising and model-error-insensitive results\cite{xiang2023learning}. 
Earlier study\cite{grady2022visual} designs a CNN network to estimate the external forces acting on two kinds of soft grippers by a monocular image. This architecture is effective for 2D but not directly transferable to 3D cases. Another study\cite{xiang2023learning} uses Long short-term memory(LSTM) to create a mapping from robot configuration information with the integration of actuation signal to the magnitude of tip force, improving the accuracy compared to the case without actuation information. The efficacy of integrating actuation information was further proven by fusing monocular images with robot actuation and state information\cite{marban2019recurrent}. Nevertheless, the paper does not concentrate on the effective feature fusion of multi-dimensional data, which is considered one of the most vital procedures that can further optimize the task\cite{zhao2019m2det}. 
Inspired by the idea of fusing multi-dimension data to improve accuracy, this paper aims to identify a method for fusing image features with actuation information and implement it into the soft robot prototype. 

This paper proposes an approach to estimate the 3D tip force of the cable-driven soft robot with monocular images, from which the essential features are extracted and preprocessed(e.g., depth image by monocular depth estimation, segment image by semantic segmentation). To deal with the dimension mismatch of high-dimension image feature vectors and low-dimension actuation information, we propose the unified feature representation module to fuse accessible actuation signals with the monocular image. Considering that the current state of the soft robot is related to the state at previous moments, this paper introduces LSTM\cite{greff2016lstm}.



Our contributions can be summarized as follows:

\begin{itemize}

\item[1)] An end-to-end method is proposed to estimate the 3D tip force of a soft robot from monocular images without requiring artificial markers. Our method is evaluated on a cable-driven soft robot and achieves the satisfying accuracy of 0.84\% of the applied forces.


\item[2)] The 2D-3D feature fusion module is designed to reduce the dimensional gap between monocular image input and 3D force output. A feature elevation mechanism is proposed to cope with low-dimension actuation features, leading to a unified feature representation fused with multi-dimensional data and providing a prerequisite for the estimation mechanism. Followed by the LSTM framework aiming to deal with time series data. 


\end{itemize}



\vspace{-1mm}
\section{METHOD}
\begin{figure*}[t]
\centering
\vspace{0mm}
\resizebox{1.0\textwidth}{!}
{
\includegraphics[scale=1.0]{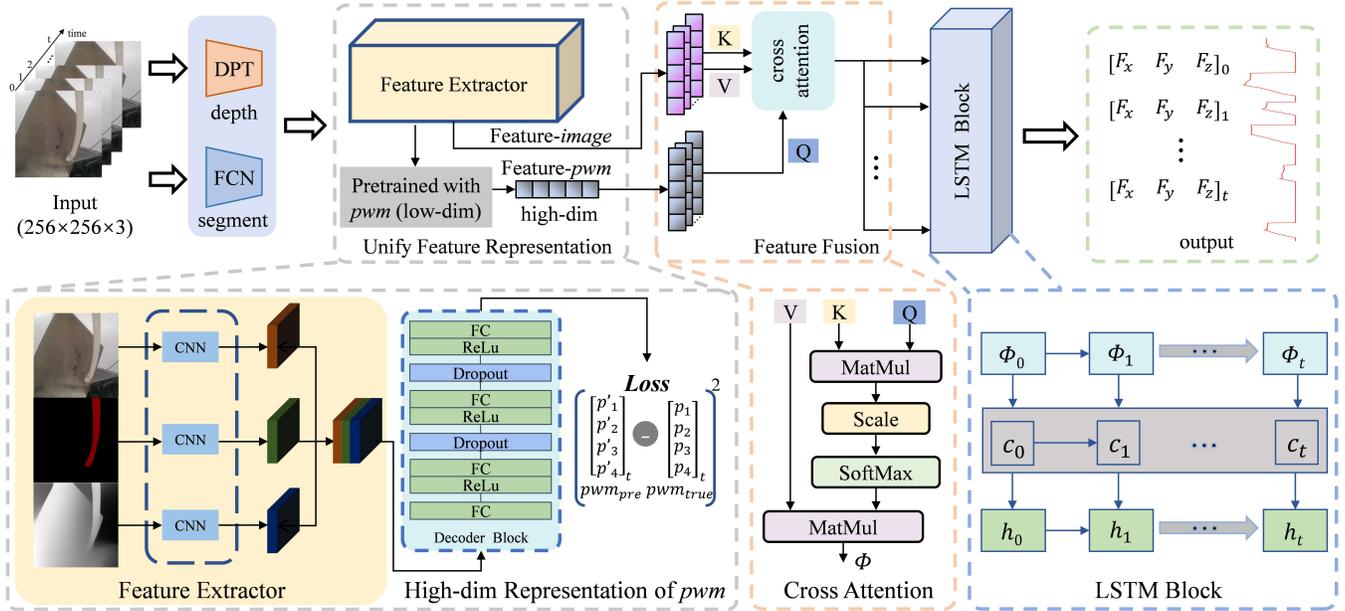}}
\vspace{-7mm}
\caption{Framework of proposed end-to-end tip force estimation network. Feature Extractor: Depth and segment images are generated from the input RGB using pre-trained DPT and FCN. All three images go through the CNNs and then flattened into image features. Unify Feature Representation: Pretrain a Feature Extractor for Feature-$pwm$ based on the PWM signal. Cross Attention: Establish a relationship between the Feature-$pwm$ and Feature-$image$ and fuse them into feature vector $\Phi$. LSTM Block: The force $F$ at time t is estimated by the LSTM network based on the time series feature vector. }
\vspace{-7mm}
\label{method_fig}
\end{figure*}

\vspace{-1mm}
Monocular camera systems offer easier deployment and are considered more suitable than stereo vision systems in certain applications. However, some technical challenges persists, hindering its application in 3D force estimation. First, the lack of an effective method to bridge the dimensional gap between 2D images and 3D force vectors.
Moreover, although early studies have demonstrated that estimation accuracy can be improved by integrating actuation inputs into the network, appropriately unifying the feature representations between different inputs remains an open question\cite{marban2019recurrent}. A similar representation form can help lightweight algorithms better understand the relationships between different inputs.
Then, discrete network architecture can not address the acknowledged hysteresis problem of the soft robot. 
Addressing the aforementioned issues, this paper develops an end-to-end algorithm to estimate 3D force vectors from monocular images and accessible actuation information.
Note that the architecture is unlimited to the form of actuation, allowing the entire algorithm to be easily applied to soft robots with diverse actuation methods.  
In the following sections, we will elaborate on the solutions to these challenges. The overall structure of our network is illustrated in Fig. \ref{method_fig}. 


\vspace{0.5mm}
\subsection{2D-3D Feature Fusion Module}


In order to make the network predict the force from 2D images.
Depth information is a prerequisite to addressing the challenge of representing soft robot 3D deflection. 
In this regard, such a problem is modeled as an image regression task which can be solved by training the network with the Mean Squared Error (MSE) loss (\ref{mse_equation}).


\vspace{-3mm}
\begin{equation}
    MSE\ =\ \frac{1}{N}\sum_{i=1}^N{\left( \boldsymbol{F}_{\boldsymbol{i}}\ -\ \boldsymbol{F}_{\boldsymbol{i}}^{\boldsymbol{'}} \right) ^2} .
    \label{mse_equation}
\vspace{-2mm}
\end{equation}
where $\boldsymbol{F}_{\boldsymbol{i}}$ represents the true value measured by a force sensor, and $\boldsymbol{F}_{\boldsymbol{i}}^{\boldsymbol{'}}$ represents the force predicted by a network.

Without loss of generality, we choose AlexNet\cite{krizhevsky2012imagenet} as the backbone network to validate the feasibility of the proposed solution. 
Although the images captured by a monocular camera contain pixel-level deflection information of the soft robot, there is still a dimension gap between the input and output, as the depth information ($z$-axis in Fig.~\ref{introduction_fig}) of force vectors cannot be reflected in the 2D images. 
Specifically, this dimension gap brings the inability to capture similar deflections that occur at different positions along the $z$-axis of the soft robot in the images. 
Note that the image plane can be considered as the $x'y'$-plane (in Fig.~\ref{introduction_fig}). There is a transformation relationship between the image space and the coordinate system of the force sensor. This type of 3D variation cannot be fully reflected in 2D images alone.

To address this limitation, we incorporate lightweight depth estimation and segmentation models by labeling parts of the images. As illustrated in Figure \ref{method_fig} (Feature Extractor), the depth estimation, which introduces pseudo-3D information, and the segmentation maps, which assist the network in focusing on the deformations of the soft robot, are both fed into the shared feature extractor. The design of the feature extractor is inspired by the Siamese network outlined, where three parallel branches share the same structure and weights.  
Then, the fused features are fed into the decoder. 
Feature fusion mechanism is designed to establish interactive relationships between different features that have the same representation, thereby ensuring better decoding.

\vspace{0.5mm}
\subsection{Unified Feature Representation}
It is known that the deflection of the soft robot is a function of actuation and external forces. The actuation information captures certain aspects of the soft robot's three-dimensional configuration. According to previous research\cite{marban2019recurrent}, combining actuation unit data (in this study, the actuation units are motors, so the actuation unit data specifically refers to the PWM signals of each motor) can reduce the uncertainty in force estimation and thus improve performance. From another perspective, PWM signals also contain information about the 3D deflection of the soft robot caused by the motors. 
By leveraging the intrinsic relationship between the configuration and actuation inputs, we strive to devise a feasible method for integrating the actuation information into image features. This endeavor aims to further enhance the accuracy of force estimation.

Given the disparity in the dimensionality between the actuation information (4$\times$1) and the image features (4096$\times$1), a direct fusion of these two distinct data types is not feasible.
It is crucial to find a unified feature representation between PWM signals and image inputs. To this end, we propose an approach, as shown in Fig. \ref{method_fig} (Unify Feature Representation). 
By training the network using the same encoder-decoder architecture, keeping the same inputs but changing the outputs to PWM signals, we can find the representation of PWM features in the image.
The processed Feature-$pwm$ (4096×1) is used as conditional information to constrain the force estimation in the decoder.
The introduction of cross-attention \cite{chen2021crossvit} aims to establish a relationship between the PWM features and the corresponding regions in the image that exhibit deformation.
By using the PWM features as queries (Q) and the image features as keys (K) and values (V), the module can focus on specific regions of the image that are influenced by the PWM signals and capture the deformations of the soft robot. This approach facilitates the fusion of information by leveraging the strengths of both modalities. 
The unified image features and PWM features have the same feature dimension and similar data representation, breaking the limitation of the learning ability of the network with different types of data.

\subsection{Sequential Multi-Feature Decoder}

Due to the acknowledged hysteresis phenomenon, the current state of the soft robot is closely correlated with its previous states, making it difficult to accurately estimate the continuously changing force based on the static monocular image merely captured at one time instant. Hence, this challenge presents a nearly insurmountable obstacle for time-discrete networks, which could be a potential explanation for why a time series network is the preferred option.
To meet the requirement of accurate and efficient prediction in physical experiments and the removal of outlier predictions, we choose the LSTM network to build a time-series force estimation framework. As shown in Fig. \ref{method_fig} (LSTM Block), the fused features are flattened and normalized as a sample input to the time-series decoder. 
Then, consecutive several samples are packaged into a batch and fed into the LSTM for time-series force prediction. By combining these components, we achieve a comprehensive network architecture. To estimate the contribution of each module to the network, we conducted a series of ablation experiments.


\section{EXPERIMENTS}

In this section, the proposed 3D force estimation method is evaluated on a cable-driven soft robot. The presented network was experimentally validated with a series of ablation experiments evaluating the contribution of each module of the proposed network. Comparison experiments were also conducted to demonstrate the performance.  

\begin{figure}[!t]
\centerline{\includegraphics[width=\columnwidth]{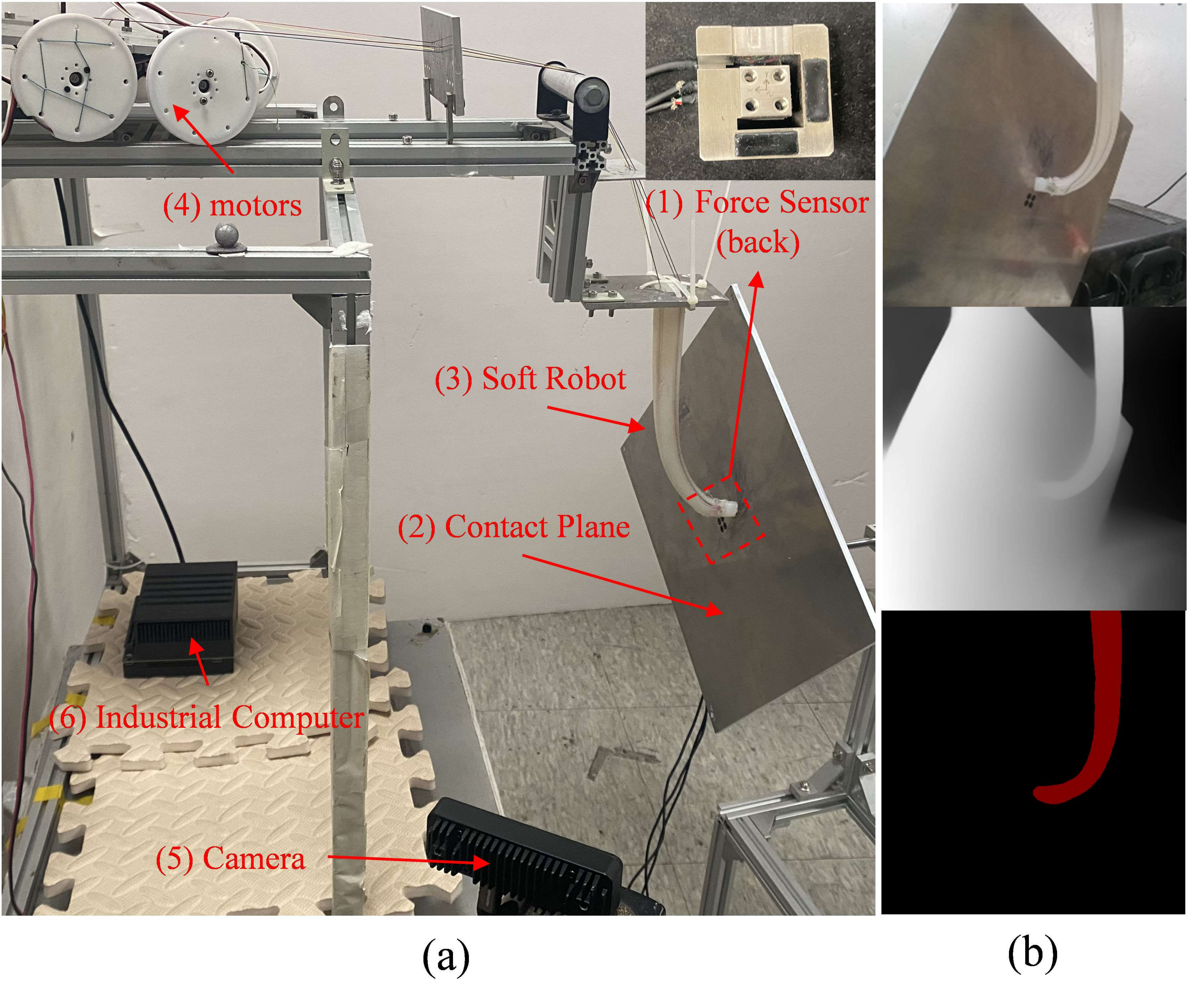}}
\vspace{-5mm}
\caption{ (a): The experimental setup for data collection. The force sensor (1) is fixed on the back (outlined) of the contact plane (2). A cable-driven soft robot (3) can contact the plane on its tip under the action of 4 motors (4). An external camera (5) captures monocular images. All of the data are saved in an industrial computer (6).  (b): Dataset collecting to estimate the tip force. From top to bottom are monocular, depth, and segment images. }
\label{device_fig}
\vspace{-5mm}
\end{figure}

\subsection{Experiment Setup}

The proposed method has been tested on a four-cable-driven soft robot platform. The whole body of the soft robot is made of silicone, with the tip capable of exerting a maximum force of 1N. Four individual motors are driven by PWM signals and actively change the cable lengths. The whole experiment setup is shown in Fig. \ref{device_fig}(a).  
PWM signals are collected as actuation information. To obtain the actual force, we use a three-dimensional force sensor with measuring frequency over 1000Hz(ZNSW-F). A fixed uncalibrated oak camera(OAK-D-Pro-W) is utilized to capture the interactive scenarios at 60Hz and 12MP resolution. 

\subsection{Data Capture}

In the process of data capture, 4 motors are driven to the randomly generated sample points at a constant speed separately. Image and force data are continuously saved during the sampling period. The sampling frequency is set to be 5Hz. We recorded 4000 sets of motion data sequences. 

Each sequence consists of dozens of images and comparable lengths of actuation and force data. Then, we use pre-trained DPT and FCN to generate the depth estimation and segment image. An example of our dataset is shown in Fig. \ref{device_fig}(b).  We remove 20\% of the data(800 sets) to create a test set for the network evaluation. The dataset is available at https://github.com/IRMVLab/Dataset-Soft.git.

\subsection{Ablations and Analysis}
The evaluation metric for this task can be described using the Root Mean Square Error ($RMSE=\sqrt{MSE}$). 
Additionally, with various fabrication methods and materials, soft robots show significant differences in load capacity. Therefore, an additional metric Mean Relative Error (MRE) with maximum contact force ${F}_{\boldsymbol{max}}$ is shown in (\ref{mre_equation}).
\begin{equation}
\vspace{-1mm}
    MRE=\frac{1}{N}\sum_{i=1}^N{\left| \frac{\boldsymbol{F}_{\boldsymbol{i}}-\boldsymbol{F}_{\boldsymbol{i}}^{\boldsymbol{'}}}{\boldsymbol{F}_{\boldsymbol{max}}} \right|} .
    \label{mre_equation}
\end{equation}

\renewcommand{\arraystretch}{1.0} 
\begin{table*}[tp]
	\centering
	\fontsize{7.5}{11}\selectfont
	\begin{threeparttable}
		\caption{Results of ablation experiments. Classify based on different Input Types (\textcolor{green}{\ding{52}} for presence, \textcolor{red}{\ding{56}} for absence; \textcolor{yellow}{\ding{52}} for arbitrary concatenation). TMCAlexPWMNet achieves the best RMSE and MRE among all methods.}
		\label{tab:perfor}
		\begin{tabular}{c|ccc|cccccccc}
			\toprule
			\multirow{2}{*}{Methods}      
            & \multicolumn{3}{c|}{Input Types} & \multicolumn{4}{c}{RMSE $\downarrow$} & \multicolumn{4}{c}{MRE (\%) $\downarrow$} \\
            \cmidrule{2-4} \cmidrule{5-8} \cmidrule{9-12}
            & Muti-Inputs & PWM-Info & Time-Series & Fx & Fy & Fz & |Fc| & Fx & Fy & Fz & |Fc| \\
            
			\midrule
			\multirow{1}*{SCAlexNet}
			& \textcolor{red}{\ding{56}} & \textcolor{red}{\ding{56}} & \textcolor{red}{\ding{56}}
            &0.017&0.025&0.0150&0.030&1.97&3.15&2.95&2.78\cr
			
            \multirow{1}*{MCAlexNet}
            & \textcolor{green}{\ding{52}} & \textcolor{red}{\ding{56}} & \textcolor{red}{\ding{56}}
            &0.016&0.018&0.014&0.024&1.71&1.89&2.36&2.06\cr
            
			\multirow{1}*{MCAlexNet-PWM}
            & \textcolor{green}{\ding{52}} & \textcolor{yellow}{\ding{52}} & \textcolor{red}{\ding{56}}
            &0.042&0.068&0.027&0.081&5.41&8.10&5.20&7.93\cr

            \multirow{1}*{MCAlexPWMNet}
            & \textcolor{green}{\ding{52}} & \textcolor{green}{\ding{52}} & \textcolor{red}{\ding{56}}
            &0.010&0.010&0.007&0.012&1.15&1.07&1.44&1.10\cr

            \multirow{1}*{TMCAlexPWMNet}
            & \textcolor{green}{\ding{52}} & \textcolor{green}{\ding{52}} & \textcolor{green}{\ding{52}}
            &{\bf 0.008}&{\bf 0.008}&{\bf 0.007}&{\bf 0.009}&{\bf 0.97}&{\bf 0.88}&{\bf 1.36}&{\bf 0.84}\cr
			\bottomrule
		\end{tabular}
	\end{threeparttable}
    \label{table_evaluation}
    \vspace{-3mm}
\end{table*}

\begin{figure}[!t]
\centerline{\includegraphics[width=\columnwidth]{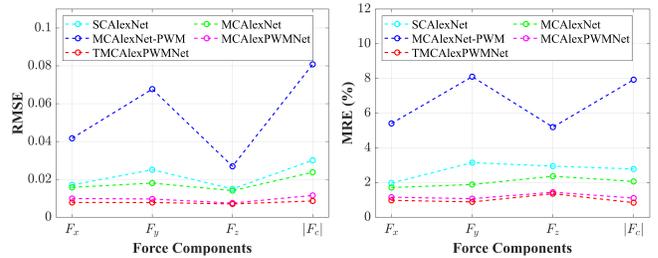}}
\vspace{-2mm}
\caption{A comparison of all proposed approaches. \textbf{Left}: The root mean square error(RMSE) of the force estimation for each axis($x$, $y$, and $z$), and $|F_c|$ is the resultant force. \textbf{right}: The mean relative error(MRE) of the force estimation for each axis($x$, $y$, and $z$), and $|F_c|$ is the resultant force. }
\vspace{-7mm}
\label{fig_error}
\end{figure}

\begin{figure*}[t]
\centering
\vspace{0mm}
\resizebox{1.0\textwidth}{!}
{
\includegraphics[scale=1.0]{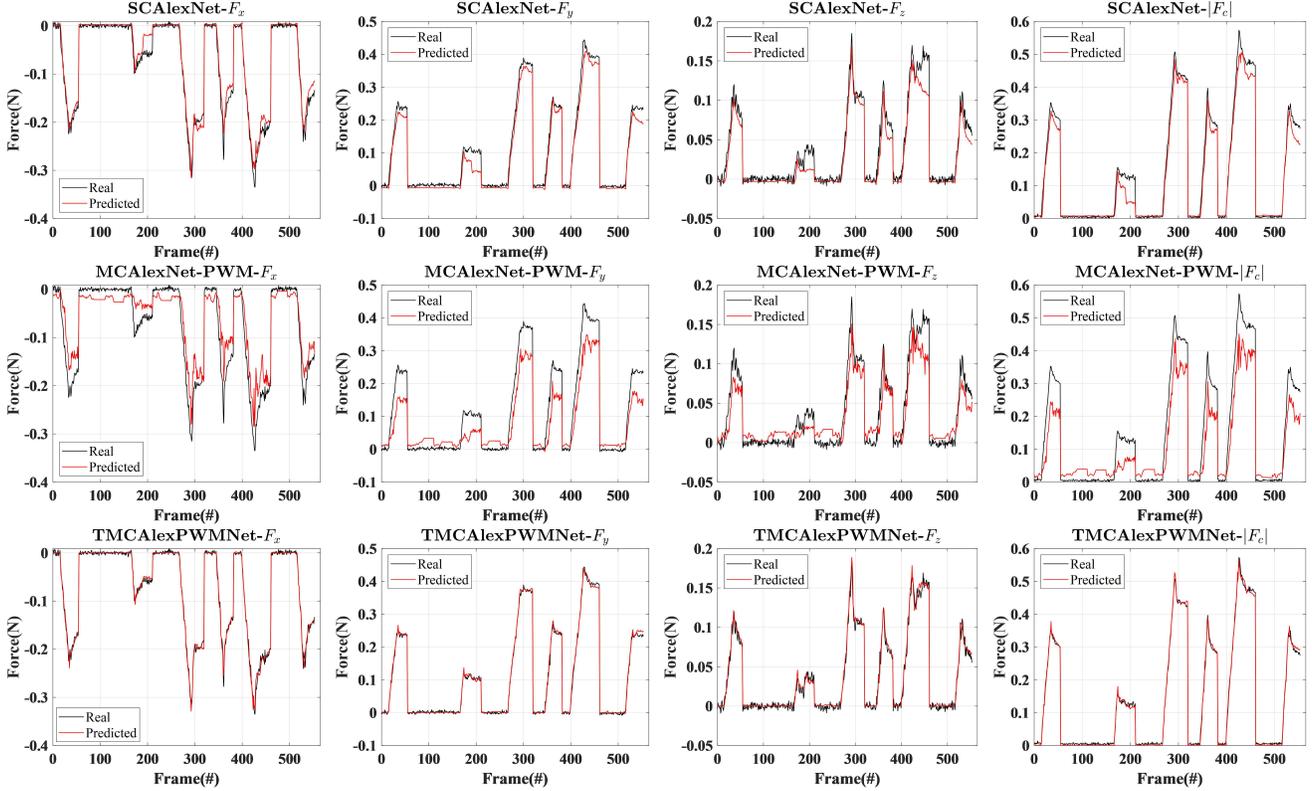}}
\vspace{-7mm}
\caption{A continuous visualization of force predictions. The first three columns show the discrepancies between the estimated and true values for each axis ($x$, $y$, and $z$), while the last column represents the resultant force. TMCAlexPWMNet performs better in all cases, particularly in terms of the moments when the soft robot makes contact with and detaches from the target.}
\vspace{-3mm}
\label{fig_visualization}
\end{figure*}

Ablation experiments build upon the design approach outlined in Section \uppercase\expandafter{\romannumeral3}. Using the same training parameters, we trained each model for 300 epochs on 80\% of the dataset. 
The specific experimental results are summarized in Table \uppercase\expandafter{\romannumeral1}. We also compile the results of all approaches into line charts, as shown in Fig. \ref{fig_error}. From our baseline network, SCAlexNet (Single-Channel estimation network), each addition of new information and improvement leads to performance enhancements. However, due to the mismatch in data formats, the MCAlexNet-PWM (Multiple-Channel estimation network) performs worse. 
The MCAlexNet (Multiple-Channel estimation network concat PWM), which has similar feature forms, overcomes this limitation and performs exceptionally well. Finally, TMCAlexPWMNet (Time-series Multiple-Channel estimation network with PWM feature), which incorporates historical information and PWM features, achieves the best performance across all metrics.

Furthermore, we visualize the experimental data in Fig. \ref{fig_visualization}, providing a detailed depiction of the differences between the predicted force and ground truth at each moment. 
It is worth noting the performance of SCAlexNet, MCAlexNet-PWM, and TMCAlexPWMNet. As our baseline for the entire work, SCAlexNet demonstrates its feasibility and surpasses our expectations. 
MCAlexNet-PWM and MCAlexPWMNet correspond to two different ways of incorporating conditional information. MCAlexNet-PWM is undoubtedly an unsuccessful attempt which prompts us to consider how to unify the representation of different forms of input. The poor performance of MCAlexNet-PWM reflects the inability to adapt to features with mismatched amplitudes and dimensions. The total force prediction MRE of 7.93\% indicates that this fusion method not only failed to improve the baseline but also resulted in worse performance. On the other hand, MCAlexPWMNet fully demonstrates the effectiveness of our idea, with a total MRE result of 1.10\%. Finally, we considered incorporating additional prior information for force estimation tasks from temporal sequences. The TMCAlexPWMNet, which incorporates an LSTM structure, exhibited significant improvements in quantitative metrics, with total MRE reduced to 0.84\% (the only structure that can reduce the MRE to below 1\%). Compared to time-discrete networks, TMCAlexPWMNet provides more accurate predictions during the moments of soft robot contact and detachment from the target.

\begin{table*}[t]\small
    \centering
    \caption{Accuracy comparison of tip force estimation on soft robots( - for absence). Works in Group {\upshape (a)} are evaluated on an elastic rod without actuation. Works in Group {\upshape (b)} are evaluated on a soft robot with actuation.}
    \vspace{-3mm}
    \fontsize{8}{11}\selectfont       
    \renewcommand{\arraystretch}{0.8}   
    \begin{tabular}{c|cccccc}
        \toprule
        \multirow{1}{*}{} & \multirow{1}{*}{Contributors} & \multirow{1}{*}{Method} & \multirow{1}{*}{Sensor} & \multirow{1}{*}{Deflection} & \multirow{1}{*}{Mean relative error}
        \cr
		\midrule

        \multirow{3}{*}{(a)}
        & Aloi et al.\cite{aloi2019estimating} & Deflection-based & External vision markers & 3D & 6.7\%
        \\
        & Al-Ahmad et al.\cite{al2021fbg} & Deflection-based & FBGs & 3D & 5.5\%
        \\
        & Diezinger et al.\cite{diezinger20223d} & Deflection-based & Stereo-vision & 3D & 2.2\%
        \cr
        \midrule
        
        \multirow{10}{*}{(b)}
        & Back et al.\cite{back2015catheter} & Deflection-based & External vision markers & 2D & 10.5\%
        \\
        & Black et al.\cite{black2017parallel} & Actuation-based & Actuator positions and forces & 3D & 8\%
        \\
        & Khan et al.\cite{khan2017force} & Deflection-based & FBGs & 3D & 6.9\%
        \\
        & Hooshiar et al.\cite{hooshiar2020accurate} & Deflection-based & External vision markers & 2D & 7.5\%
        \\
        & Yasin et al.\cite{yasin2021joint} & Actuation-based & Actuator forces & 3D & 9.55\%
        \\
        & Aloi et al.\cite{aloi2022estimating} & Deflection-based & External vision markers & 3D & -
        \\
        & Diezinger et al.\cite{diezinger2023trirod} & Deflection-based & Binocular-vision & 3D & 10.5\%
        \\
        & Brumfiel et al.\cite{brumfiel2024image} & Deflection-based & Monocular-vision & 2D & 7\%
        \\
        & Proposed method & Deflection-based & Monocular-vision & 3D & \bf{0.84\%}
        \cr
        \bottomrule
        
    \end{tabular}
    \label{table_comparison}
    \vspace{-6mm}	
\end{table*}

\subsection{Comparison}
We present a comparative analysis of our results alongside those of other related works, as outlined in Table \ref{table_comparison}. The previous investigations within group (a) evaluate their force estimation methodologies on an elastic rod without actuation. For instance, the previous study\cite{aloi2019estimating} prearranges external vision markers on the elastic rod for the discrete shape measurements, optimizing the difference between real and theoretical shapes to derive external force estimations. This approach is extended to \cite{aloi2022estimating} which considers actuation effects. Their method achieves 6.7\% accuracy, but its slow convergence, taking nearly 2.1 seconds, limits its real-time application. Another study\cite{al2021fbg} employs FBGs to obtain discretized curvature measurements along the rod's length and proposes two force estimation methods based on piecewise polynomial curvature segmentation or Cosserat rod theory. The evaluation on a Nitinol rod embedded with a multi-core fiber shows a mean relative error of 5.5\%. \cite{diezinger20223d} reconstructs the 3D shape of the elastic rod using stereo-vision and directly estimates tip forces by solving a series of balance equations. Their method can run in real-time and achieves the best-reported accuracy of 2.2\%. 

Due to the absence of actuation influence, group (a) generally attains superior accuracy in force estimation compared to group (b), which evaluates their methods on the soft robot with actuation. Actuation-based methods\cite{black2017parallel},\cite{yasin2021joint} estimate the distal forces through actuation information(actuator positions or forces), with a mean relative error of less than 10\%. Their approaches prove beneficial in situations where only actuation information is available. \cite{back2015catheter} employs a deflection-based method, mounting 12 vision markers onto a soft robot in advance and estimating planar contact forces with RGB images via a simplified Cosserat rod model. Their approach is evaluated on a 4-tendon-driven catheter and achieves an accuracy of 10.5\%. Previous work \cite{khan2017force} leverages FBG sensors to measure strain in a 4-tendon-driven soft manipulator, observing a mean absolute error of 11.2mN (6.9\% of applied magnitude). \cite{diezinger2023trirod} estimates the unknown external load by optimizing the difference in robot shape taken by virtual cameras and real cameras(binocular vision). This method is tested on a continuum parallel robot(called TriRod) and shows a mean relative error of 10.5\% for the estimation of loads up to 633 mN. Study\cite{brumfiel2024image} estimates the contact forces from monocular images with an accuracy of 7\%. This work represents the initial trial of utilizing a monocular camera for contact force estimation of a soft robot. Note that their method can not solve 3D deflection cases. Our proposed method reduces the mean relative error of soft robot force estimation to less than 1\%(specifically 0.84\%) for the first time. By utilizing monocular images as input, our method delivers 3D force output and effectively handles 3D deflection scenarios. 

\section{CONCLUSIONS}

In this work, an end-to-end network is introduced for estimating the 3D tip force of the soft robot with monocular images. The 2D-3D feature fusion module aims to minimize the dimensional disparity between monocular image input and 3D force output. 
Besides, to combine the high-dimension image feature with the low-dimension actuation information (PWM signals), we also introduce an auxiliary task to transfer the actuation information into the image feature to provide similar feature forms. Then LSTM is introduced to deal with time series data. The proposed method has experimentally validated on the cable-driven soft robot prototype and demonstrated the force estimation accuracy of 0.84\%(MRE). 
We aim to estimate the body forces along the soft robot from monocular images in the future.

\addtolength{\textheight}{-6cm}   
\bibliographystyle{ieeetr}
\bibliography{myref}

\end{CJK}
\end{document}